\documentclass[times,twocolumn,final,authoryear,5p]{elsarticle}

\usepackage{amssymb}
\usepackage{amsmath}

\usepackage{hyperref}

\makeatletter
\def\ps@pprintTitle{%
    \let\@oddhead\@empty
    \let\@evenhead\@empty
    \def\@oddfoot{\footnotesize\itshape
         {Computer Vision and Image Understanding (\url{https://doi.org/10.1016/j.cviu.2022.103582}) } \hfill}%
    \let\@evenfoot\@oddfoot
    }
\makeatother

\usepackage[labelformat=simple]{subcaption}

\usepackage{tabularx}
\usepackage{booktabs}
\usepackage{makecell}
\usepackage{csquotes}

\newcommand{\revisedsection}[1]{{#1}}

\usepackage[ruled, vlined, noend]{algorithm2e}

\newcolumntype{b}{X}
\newcolumntype{s}{>{\centering\arraybackslash\hsize=.089\hsize}X}
\newcolumntype{u}{>{\centering\arraybackslash\hsize=.17\hsize}X}
\newcolumntype{t}{>{\centering\arraybackslash\hsize=.4\hsize}X}
\newcolumntype{v}{>{\centering\arraybackslash\hsize=.2\hsize}X}

\begin{document}

\begin{frontmatter}

\title{Balanced softmax cross-entropy for incremental learning with and without memory}

\author[label1,label2]{Quentin Jodelet\corref{cor1}} 
\ead{jodelet@net.c.titech.ac.jp}
\author[label2]{Xin Liu}\ead{xin.liu@aist.go.jp}
\author[label1,label2]{Tsuyoshi Murata}\ead{murata@c.titech.ac.jp}

\address[label1]{Department of Computer Science, Tokyo Institute of Technology, 2-12-1-W8-59 Ookayama, Meguro, 152-8552, Tokyo, Japan}
\address[label2]{Artificial Intelligence Research Center, AIST, 2-4-7 Aomi, Koto, 135-0064, Tokyo, Japan}

\cortext[cor1]{Corresponding author}

\begin{abstract}
When incrementally trained on new classes, deep neural networks are subject to catastrophic forgetting which leads to an extreme deterioration of their performance on the old classes while learning the new ones. Using a small memory containing few samples from past classes has shown to be an effective method to mitigate catastrophic forgetting. However, due to the limited size of the replay memory, there is a large imbalance between the number of samples for the new and the old classes in the training dataset resulting in bias in the final model. To address this issue, we propose to use the Balanced Softmax Cross-Entropy and show that it can be seamlessly combined with state-of-the-art approaches for class-incremental learning in order to improve their accuracy while also potentially decreasing the computational cost of the training procedure. We further extend this approach to the more demanding class-incremental learning without memory setting and achieve competitive results with memory-based approaches. Experiments on the challenging ImageNet, ImageNet-Subset, and CIFAR100 benchmarks with various settings demonstrate the benefits of our approach.
\end{abstract}


\end{frontmatter}


\section{Introduction}
\label{section:introduction}
In a class incremental learning scenario, the complete training dataset is not available at once. Instead, the training samples are progressively available, few classes at a time. The model has to learn new classes in a sequential manner, similarly to the human learning process. This ability to incrementally learn new knowledge is critical for various real world applications where new samples may appear sequentially and it is not possible to either store them all due to memory constraint or re-train the model from scratch as new data is encountered due to time or computational power limitations. For example, a robot learning to classify new objects while interacting with its environment or a face recognition system should be able to incrementally learn new knowledge during their lifetime without having to be re-trained on previously learned faces or objects each time a new element is encountered. Although deep neural networks achieve state-of-the-art performance across a wide spectrum of applications in computer vision, training them incrementally remains challenging. It is caused by the high propensity of deep neural networks to almost fully forget previously learned classes while learning new ones. This situation is known as catastrophic forgetting~\citep{french1999catastrophic, MCCLOSKEY1989109, parisi2019continual}. 

In this context, using a small memory buffer to store samples of previously encountered classes in order to use them for rehearsal while learning new classes has been proven by multitude of recent works to be an effective solution to mitigate catastrophic forgetting. However, as the size of the replay memory is limited in practice, a large imbalance issue appears, i.e.\ at a given time, the model will mainly be trained using data from new classes and only a few from previous classes. This leads the model to become biased towards the most recently learned classes and resulting in a deterioration of its overall performance. This phenomenon gets worse as the size of the memory decreases to the point where no memory at all is available.  Methods designed to tackle this issue usually rely on additional finetuning steps using a small balanced dataset after each incremental step, oversampling, or specifically designed classifiers.

In this work, we propose a novel approach to mitigate the bias towards new classes in large-scale class incremental learning with and without a replay memory. Our proposed method modifies the training procedure by using the Balanced Softmax activation function~\citep{NEURIPS2020_2ba61cc3} for the Cross-Entropy loss instead of the commonly used Standard Softmax function. By combining the Balanced Softmax Cross-Entropy loss with state-of-the-art approaches for class incremental learning with rehearsal memory, it is possible to further improve the average incremental accuracy of those methods on competitive benchmarks. Our method also potentially decreases the computational cost of the training procedure as the Balanced Softmax Cross-Entropy loss does not require any additional balanced finetuning step. Moreover, we propose an additional meta-learning algorithm that can further improve the accuracy of the models in some settings. Finally, we propose an extension of the Balanced Softmax Cross-Entropy, named the Relaxed Balanced Softmax Cross-Entropy, specifically designed for class incremental learning without memory. We show that this newly proposed loss can be applied to approaches initially conceived for class incremental learning with a replay memory in order to adapt them to the exemplar-free setting without further modification. They then achieve state-of-the-art results \revisedsection{and are competitive with memory-based approaches}.

To summarize, our contributions are three-fold:
\begin{itemize}
    \item We propose the use of the Balanced Softmax Cross-Entropy as a mitigation method for large-scale class incremental learning and we show that it can be seamlessly combined with advanced state-of-the-art methods in order to improve their accuracy while also decreasing the computational complexity of their training procedure.
    \item We propose a meta-learning algorithm that can improve the accuracy of the Balanced Softmax Cross-Entropy in some settings.
    \item We introduce an extension of the Balanced Softmax Cross-Entropy \revisedsection{specifically designed} for class incremental learning without memory, \revisedsection{named Relaxed Balanced Softmax Cross-Entropy}. We show that it achieves state-of-the-art performances and can efficiently balance the trade-off between stability and plasticity.
\end{itemize}

In this work, we extend our previously published conference paper \citep{10.1007/978-3-030-86340-1_31} by considering additional settings and experiments for our proposed method. First, we explore the class incremental learning without memory setting. In this setting, no exemplar of old classes can be stored in the memory, which makes it a challenging problem. \revisedsection{It is also not well-addressed by existing methods.} We highlight the limitations of the Balanced Softmax Cross-Entropy in this particularly demanding setting and propose an improved version, named Relaxed Balanced Softmax Cross-Entropy, which significantly improves the accuracy of our proposed method and achieves state-of-the-art performances. Secondly, we perform additional experiments on two more challenging scenarios for class incremental learning with replay memory: the 25 incremental steps setting on CIFAR100 and ImageNet-Subset and the 10 incremental steps setting without large base step on ImageNet as initially used by \cite{Rebuffi_2017_CVPR} and \cite{wu2019large}. Finally, we also add supplementary ablation study and figures.

\section{Related work}
\label{section:relatedWork}
In recent years, continual learning and incremental learning for deep neural networks gained more and more attention, following the success of deep learning based methods during the past decade. There exists a large variety of scenarios and settings for continual learning~\citep{hsu2018re, LOMONACO2022103635}. In our work, we will only consider the class incremental learning scenario. Most methods designed for this scenario usually combine three distinct components when applied to large scale datasets: constraints to preserve past knowledge, rehearsal memory, and bias mitigation methods. 

\cite{hinton2015distilling} initially proposed the distillation loss for transferring knowledge from a large teacher model into a smaller student model. \cite{li2017learning} adapted this method to continual learning by distilling the knowledge of the model learned during the previous step into the next step using the output logits of the models. This method was then applied by several authors~\citep{10.1007/978-3-030-01258-8_15, Rebuffi_2017_CVPR, wu2019large}.
Recently, several proposals have been made to modify and improve distillation for incremental learning. \cite{Hou_2019_CVPR} proposed a novel distillation loss named Less forget Constraint which is applied on the final class embeddings instead of the output logits. \cite{Dhar_2019} proposed the Attention Distillation Loss which penalizes changes in the attention maps of the classifiers. \cite{douillard2020podnet} proposed a new distillation loss using a pooling function and applied it to several intermediate layers of the neural network in addition to the final class embedding. \cite{10.1007/978-3-030-58529-7_16} proposed to model the class embedding topology using an elastic Hebbian graph and then used a topology-preserving loss to constrain the change of the neighboring relationships of the graph during each incremental step. Similarly, \cite{lei2020class} adopted a feature-graph preservation approach and proposed the weighted-Euclidean regularization to preserve the knowledge.
Lastly, \cite{Simon_2021_CVPR} proposed to use the geodesic flow for knowledge distillation in order to preserve past knowledge more efficiently. 

Using a small replay memory containing samples from previously encountered classes~\citep{chaudhry2019tiny} has been shown to be a simple yet effective method to mitigate catastrophic forgetting. In order to increase the number of stored samples for a fixed size, it is possible to rely on compression~\citep{caccia2020online} or to store intermediate representations~\citep{Hayes_2020} instead of the input images. It is also possible to store the output logits for distillation instead of the true labels if the model from the previous step is not available~\citep{NEURIPS2020_b704ea2c}. \cite{liu2020mnemonics} proposed to parameterize the exemplars of the memory and to learn them in an end-to-end manner through bi-level optimizations.

\begin{figure*}
   \begin{center}
          \includegraphics[width=0.75\linewidth]{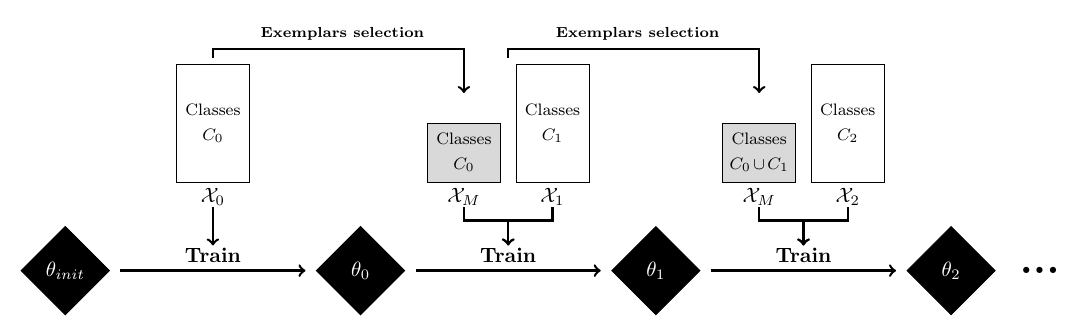}
   \end{center}
   \caption{Illustration of the class incremental training procedure. At each incremental step $i$, the model is trained with the new data $\mathcal{X}_i$ containing samples from new classes $C_i$ and the memory $\mathcal{X}_M$ containing few samples from previously encountered classes $\bigcup_{t=0}^{i-1} C_t $. The step 0, also called the base step, contains the base classes.}
\label{fig:trainingProc}
\end{figure*}

However, the limited size of the replay memory results in an unbalanced training set mainly composed of samples from the new classes. Several recent works have highlighted this problem and proposed different methods to address it. \cite{Rebuffi_2017_CVPR} introduced iCaRL which relies on a nearest-mean-of-exemplars classifier to tackle the issue. \cite{10.1007/978-3-030-01258-8_15} proposed to use more data augmentation on the training set and to finetune the model on a small balanced dataset after each incremental step.
This additional balanced finetuning step was then reused in numerous works~\citep{Hou_2019_CVPR,douillard2020podnet}.
\cite{wu2019large} and \cite{Hou_2019_CVPR} observed that the classifier part of the neural network is the main reason for the bias of the model towards the new classes. Therefore, ~\cite{wu2019large} proposed to learn a two-parameter linear model on a small balanced dataset to correct the bias of the last fully connected layer while \cite{Hou_2019_CVPR} proposed to use cosine normalization on the classifier. A recent work~\citep{Ahn_2021_ICCV} also proposed to use oversampling of old classes and to separately compute the softmax probabilities of the new and old classes for the Cross-Entropy loss.

Nevertheless, storing samples from past classes inside the replay memory may not always be possible due to privacy reasons. This situation, referred to as exemplar-free incremental learning or incremental learning without memory, significantly complicates the initial problem. Pioneering works proposed to learn a generative model~\citep{NIPS2017_0efbe980,NEURIPS2018_a57e8915} to generate samples from old classes. Recently, \cite{Yu_2020_CVPR} proposed an approach that does not rely on a generative model but which approximates the semantic drift of the embedding space using new data only while \cite{Zhu_2021_CVPR} and \cite{Wu_2021_ICCV} proposed to use self-supervised learning to learn richer representations. Concurrently, \cite{Hu_2021_CVPR} proposed a solution that aims to distill the effects of old data without having to store them in the replay memory. 

In our work, we propose to use the Balanced Softmax Cross-Entropy loss to address the problem of unbalanced training set for class incremental learning with and without storing samples from past classes in memory. Compared to previously presented methods, the Balanced Softmax Cross-Entropy loss does not require additional training steps or oversampling while reaching similar or higher accuracy on competitive benchmarks.

\section{Proposed method}
\label{section:proposedMethod}

The objective of class incremental learning is to learn a unified classifier (also denoted single-head classifier) from a sequence of training steps, each containing new and previously unseen classes, as described in Figure \ref{fig:trainingProc}. The first training step called the base step, is followed by several incremental steps numbered from 1 to $\mathcal{T}$, for a total of $\mathcal{T}+1$ training steps, each composed of the training set $\mathcal{X}_t$ containing samples from the set of classes $C_t$. Each incremental step contains different classes such that $\bigcap_{t=0}^{\mathcal{T}} C_t = \varnothing$. During each incremental step, in addition to $\mathcal{X}_t$, the model also has access to the small replay memory $\mathcal{X}_M$ containing samples from classes encountered during previous incremental steps.
The number of classes learned up to the incremental step $t$ included is denoted as $N_t$. \revisedsection{At the end of each training step, the model is evaluated on the test data of all classes seen so far $\bigcup_{t=0}^{i} C_t$ where $i$ denotes the current training step.}

\subsection{Incremental learning baseline}
To highlight the strengths of our proposed method, we use a simple baseline for incremental learning, denoted IL-baseline, initially proposed by \cite{wu2019large}. This baseline consists of a deep neural network combined with a small replay memory and optimized using two distinct losses: the Standard Softmax Cross-Entropy loss and the distillation loss. 

The total loss $\mathcal{L}$ used to train the model, is defined as a weighted sum of the distillation loss $\mathcal{L}_d$ and the Standard Softmax Cross-Entropy loss $\mathcal{L}_c$:
 \begin{equation}
 	 \mathcal{L} = \rho \mathcal{L}_d + (1-\rho) \mathcal{L}_c
 \end{equation}
 where $\rho$ is defined as $\frac{N_{t-1}}{N_{t}}$ with $N_t$ the total number of classes at incremental step $t$ and is used to balance the importance of the two losses. The importance of the distillation loss is modified based on the number of new classes that are learned during the training step compared to the total number of classes that has already been learned.

At the beginning of each incremental step $t$, the parameters from the previous step $\theta_{t-1}$ are first copied to initialize the new parameters $\theta_t$. The old parameters $\theta_{t-1}$ are then used to maintain the knowledge of previously learned classes using the distillation loss~\citep{hinton2015distilling,li2017learning}. For each training sample $(x,y) \in \mathcal{X}_t\cup \mathcal{X}_M$, the distillation loss $\mathcal{L}_d$ is defined as:
\begin{equation}
\begin{gathered}
 	 \mathcal{L}_d (x) =  \displaystyle\sum_{k=1}^{N_{t-1}}  - \hat{p}_k(x) \; \text{log}(p_k(x)) \; T^2 \;, \\
 	 \hat{p}_k(x)=\frac{ e^{\hat{z}_k(x)/T}}{\sum_{j=1}^{N_{t-1}} e^{\hat{z}_j(x) / T}}, \; \; \; p_k(x)=\frac{ e^{z_k(x)/T}}{\sum_{j=1}^{N_{t-1}} e^{z_j(x) /T }}
 	 \end{gathered}
 	 \label{eq:distillation}
\end{equation}
where $x$ is the input image, $y$ is the associated ground truth label, $z(x) = [z_1(x), ..., z_{N_t}(x)]$ is the output logits of the current model $\theta_t$,  $\hat{z}(x) = [\hat{z}_1(x), ..., \hat{z}_{N_{t-1}}(x)]$ is the output logits of the model at the previous incremental step $\theta_{t-1}$ and $T$ is the temperature.

Usually, the replay memory can be constrained by either a global budget or by the number of samples saved per class. Using a global budget means that the total memory size is fixed at the beginning of the training procedure and is equally divided among the already learned classes. As the number of already encountered classes increases, the number of samples stored for each class decreases.
On the other hand, using a fixed number of samples stored per class, also known as growing memory, means that the number of samples per class will not change during the training procedure, which means that the total size of the memory increases at each incremental step. The latter approach is usually considered to be the more challenging one.
The herding selection~\citep{10.1145/1553374.1553517} is often used to select the samples as it has been shown to be more efficient than the random selection~\citep{belouadah2020comprehensive}.

\subsection{Balanced Softmax Cross-Entropy}

\begin{figure}
\captionsetup[subfigure]{labelformat=empty}
\begin{center}
\begin{subfigure}[t]{\columnwidth}
\begin{center}
           \includegraphics[width=0.75\columnwidth]{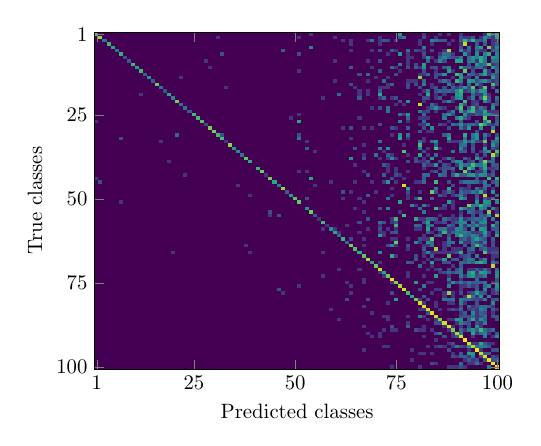}
        \caption{IL-Baseline \\ Final overall top-1 accuracy: 31.25\%} 
\end{center}

    \end{subfigure}
    \end{center}
   \caption{Confusion matrix (transformed by log(1 + x) for better visibility) of the IL-Baseline trained on CIFAR100 with a base step containing half of the classes followed by 5 incremental steps, using 20 samples per class stored in the replay memory. Figure best viewed in color.}
\label{fig:confmat_ILbaseline}
\end{figure}

Due to the limited size of the replay memory, the training set $\mathcal{X}_t\cup \mathcal{X}_M$ contains only few tens of samples for each of the old classes while containing hundreds or thousands of samples for each of the new classes at each incremental step. The testing set, however, contains the same number of samples for both the new classes and the old classes. This discrepancy induces a bias towards the most recently learned classes~\citep{belouadah2019il2m, Hou_2019_CVPR,wu2019large} as illustrated in Figure~\ref{fig:confmat_ILbaseline} with the IL-Baseline procedure. As can be seen from the figure, the model tends to predict the classes which had the largest number of samples in the training set during the last incremental steps (the new classes) rather than the old classes. This situation is similar to the Long-Tailed Visual Recognition problem where a model is evaluated on a balanced test dataset after being trained on a dataset composed of a few classes that are over-represented (the head classes) and a large number of classes that are under-represented (the tail classes).

Based on this observation, we propose to replace the Standard Softmax activation function with the Balanced Softmax for the Cross-Entropy loss during the training procedure. This activation function has been initially introduced by \cite{NEURIPS2020_2ba61cc3} to address the label distribution shift between the training set and test set in Long-Tailed Visual Recognition.
 
The Balanced Softmax activation function is defined as:
 
\begin{equation}
 	 q_k(x)=\frac{ \lambda_k  e^{z_k(x)}}{\sum_{j=1}^{N_t} \lambda_j e^{z_j(x)} }  \;
 	 \text{ with } \lambda_i = n_i
 \label{eq:balancedSoftmax}
 \end{equation}
where $x$ is the input image, $z(x) = [z_1(x), ..., z_{N_t}(x)]$ is the output logits of the current model $\theta_t$, $n_i$ is the number of samples in the training set for the $i$\textsuperscript{th} class \revisedsection{for the current step $t$ and $N_t$ is the number of classes learned up to the current incremental step $t$ included.} \revisedsection{For detailed analysis of the Balanced Softmax and its relation to the general class imbalance problem, we refer readers to \cite{NEURIPS2020_2ba61cc3}.}

The new classification loss $\mathcal{L}_c$ used during the training procedure is then defined as the Cross-Entropy loss using the Balanced Softmax instead of the Standard Softmax:
\begin{equation}
 	\mathcal{L}_c (x)= \sum_{k=1}^{N_t} -\; \delta_{k=y} \; \text{log}(q_k(x))
 \label{eq:balancedCE}
 \end{equation}
 where $\delta$ is the indicator function and $y$ is the associated ground truth label to $x$.

The Balanced Softmax Cross-Entropy loss, denoted BalancedS-CE, can be used as a replacement for the Standard Softmax Cross-Entropy loss in the previously defined IL-Baseline or in any other model for incremental learning. It should be noted that the Balanced Softmax Cross-Entropy does not increase the computational cost of the training procedure in any notable way compared to the Standard Softmax Cross-Entropy. 

\subsection{Meta Balanced Softmax Cross-Entropy}
The expression of the Balanced Softmax presented in Equation~(\ref{eq:balancedSoftmax}) allows for direct control of the importance of each class by selecting a dedicated weighting coefficient $\lambda_i$ for each of them, which may be different from the number of samples for this class in the training dataset, similarly to~\cite{khan2017cost}. In the context of large scale incremental learning, the modification of these weighting coefficients offers a new method for controlling the plasticity-rigidity trade-off of the trained model and also for controlling separately the importance of each individual class. 

We propose to extend the Balanced Softmax by introducing a new global weighting coefficient $\alpha$ to control the importance of the past classes:

\begin{equation}
 	 q_k(x)=\frac{ \lambda_k  e^{z_k(x)}}{\sum_{j=1}^{N_t} \lambda_j e^{z_j(x)} } \;
 	 \text{ with } \lambda_i =
    \begin{cases}
        \alpha \; n_i,& \text{if } i\in P\\
        n_i, & \text{otherwise}
    \end{cases}
 \end{equation}
 where $P$ is the set of previously encountered classes and the weighting coefficient $\alpha$ is a real number, usually between 0 and 1.

The expression of the $\lambda_i$ coefficient for the new classes does not change, but for the old classes, the number of images in the training set $n_i$ is multiplied by the weighting coefficient $\alpha$.
By decreasing $\alpha$, it is possible to increase the importance of the old classes and reduce the plasticity of the model depending on the specific requirements of the application. This new expression of the Balanced Softmax is equivalent to the one described in Equation~(\ref{eq:balancedSoftmax}) for $\alpha$ equal to 1.0.

In practice, it appears that 1.0 may not be the optimal value for $\alpha$ when only considering the average incremental accuracy of the model. However, it is difficult to determine beforehand a satisfying value for $\alpha$ without performing several trials with different values. Therefore, to further improve the accuracy of the Balanced Softmax Cross-Entropy for Incremental Learning, we propose a new training procedure, named Meta Balanced Softmax Cross-Entropy (Meta BalancedS-CE), in order to slightly adjust the weighting coefficient $\alpha$ of the Balanced Softmax during the training as described by Algorithm~\ref{alg:metaBalancedSoftmax}. 

\begin{algorithm}
\small
\SetKwInOut{Input}{input}
\SetKwComment{Comment}{$\triangleright$\ }{}
\SetAlgoLined
\DontPrintSemicolon
 Initialize parameters $\theta$ and function $Z$ of the model\;
 Initialize memory $\mathcal{X}_M$\;
\For{$t\gets1$ \KwTo $\mathcal{T}$}{
 $\alpha \leftarrow 1.0$\;
 $D_t, B_t \leftarrow $ split($\mathcal{X}_t\cup \mathcal{X}_M$) \;
 \For{$r\gets1$ \KwTo $R$}{
 \For{$(X,Y) \sim D_t$}{
  $\theta^* \leftarrow \theta - \triangledown_{\theta} $balancedLoss($Z(\theta, X)$, $ Y )$\;
 $(\bar{X}, \bar{Y}) \sim B_t$ \;
 $\alpha \leftarrow \alpha - \triangledown_{\alpha} $softmax\_CE($Z( \theta^*, \bar{X})$, $ \bar{Y} )$\;

 $\theta \leftarrow \theta - \triangledown_{\theta} $balancedLoss($Z(\theta, X)$, $ Y )$ \;

 }
 }
  $\mathcal{X}_M \leftarrow$ updateMemory($\mathcal{X}_t$)\;
 }
 \caption{Meta Balanced Softmax Cross-Entropy training procedure}
 \label{alg:metaBalancedSoftmax}
 
\end{algorithm}

Instead of using the same fixed weighting coefficient $\alpha$ during the complete training procedure, we propose to jointly learn $\alpha$ during the training of the deep neural network. To achieve this, we propose a meta-learning algorithm that estimates at each optimization step the optimal value of $\alpha$ using a balanced validation set $B_t$. At the beginning of each incremental step, the unbalanced training set $\mathcal{X}_t\cup \mathcal{X}_M$ composed of the samples from the new classes and the samples of old classes stored in the memory is split into a training set $D_t$ and a validation set $B_t$. Unlike $D_t$ which is a large unbalanced dataset, $B_t$ is a smaller set containing the same number of samples for every class.

At each optimization step, a temporary model $\theta^*$ is created by training the current model $\theta$ on the incoming batch of data $(X,Y)$ from $D_t$ using the balanced loss which is the sum of the Balanced Softmax Cross-Entropy loss and secondary losses (such as the distillation loss). By using a batch $(\bar{X}, \bar{Y})$ from the balanced validation set $B_t$, the value of $\alpha$ is then updated using the gradient of the standard Softmax Cross-Entropy loss of $(Z(\theta^*, \bar{X}), \bar{Y})$ with respect to $\alpha$. Finally, we update the current model $\theta$ on the batch $(X,Y)$ previously sampled from $D_t$ using the balanced loss with the newly learned value of $\alpha$.

Unlike the Balanced Softmax Cross-Entropy which does not change the computational cost of the training procedure compared to the Standard Softmax Cross-Entropy, the Meta Balanced Softmax Cross-Entropy has an impact on the training procedure. The method requires the computation of gradients through the optimization process. One of the main drawbacks is a large increase in the memory requirement which may make it more difficult to combine this method with some existing methods for incremental learning.

\begin{figure}[ht]
\begin{subfigure}[t]{\columnwidth}
        \centering
        \includegraphics[width=0.75\columnwidth]{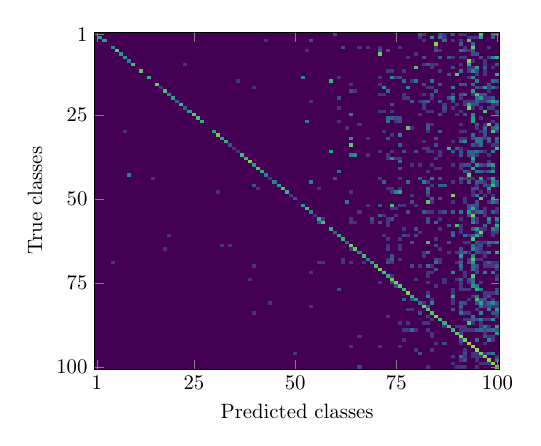} 
        \caption{LUCIR \\ Final overall top-1 accuracy: 38.36\%}
        \label{fig:confmat_balancedsoftmax_0mem:lucir}
    \end{subfigure}
    
    \vspace{\floatsep}
    
    \begin{subfigure}[t]{\columnwidth}
        \centering
        \includegraphics[width=0.75\columnwidth]{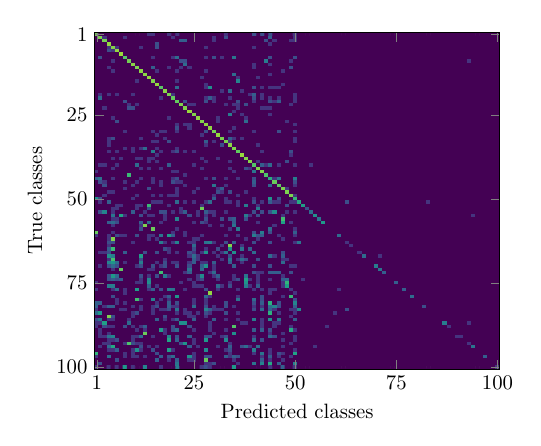} 
        \caption{LUCIR w/ Balanced Softmax Cross-Entropy  \\ Final overall top-1 accuracy: 42.00\% }
        \label{fig:confmat_balancedsoftmax_0mem:balsoft}
    \end{subfigure}
   \caption{Confusion matrix (transformed by log(1 + x) for better visibility) of LUCIR \citep{Hou_2019_CVPR} trained on ImageNet-Subset with a base step containing half of the classes followed by 5 incremental steps with 0 sample per class stored in the replay memory using \subref{fig:confmat_balancedsoftmax_0mem:lucir} the Standard Softmax Cross-Entropy and \subref{fig:confmat_balancedsoftmax_0mem:balsoft} the Balanced Softmax Cross-Entropy. Figure best viewed in color.}
\label{fig:confmat_balancedsoftmax_0mem}
\end{figure}

\subsection{Relaxed Balanced Softmax Cross-Entropy}

In the exemplar-free setting, the $\lambda_i$ coefficient for every old class will be equal to 0 in the Equation~(\ref{eq:balancedSoftmax}) as it is not possible to save any exemplar from past classes into the memory. Therefore, in this particular setting, the Balanced Softmax Cross-Entropy is equivalent to the standard Softmax Cross-Entropy applied on the new classes only, supposing that the number of images per class in the dataset is the same for all the new classes. This constraint heavily limits the plasticity of the model as shown in Figure \ref{fig:confmat_balancedsoftmax_0mem:balsoft}. In this experiment, LUCIR \citep{Hou_2019_CVPR} is trained on ImageNet-Subset with a base step containing half of the classes followed by 5 incremental steps using the Balanced Softmax Cross-Entropy. At the end of the training procedure, while achieving 79.44\% accuracy on the 50 base classes, the model only reaches 4.56\% on the latest 50 classes. Compared to the 81.16\% of accuracy on the base classes at the end of the base step, this means that during the five incremental steps following the base step, the model mostly protected its previous knowledge at the cost of not being able to adapt to the new classes. By comparison, when LUCIR is trained using the Standard Softmax Cross-Entropy, Figure \ref{fig:confmat_balancedsoftmax_0mem:lucir} shows that the opposite happens: \revisedsection{the accuracy on the 50 base classes drops from 82.12\% at the end of the base step to 29.76\% at the end the final step while achieving a similar overall final accuracy.} The model \revisedsection{trained with the Standard Softmax Cross-Entropy} is only able to correctly classify the most recent classes as the method was not designed for class incremental learning without memory. \revisedsection{A more suitable solution would therefore be in between those two extreme approaches that are the Balanced Softmax Cross-Entropy and the Standard Softmax Cross-Entropy in the zero memory setting.}

To that end, we proposed an extended version of the Balanced Softmax Cross-Entropy specifically designed for class incremental learning without memory named Relaxed Balanced Softmax Cross-Entropy (Relaxed BalancedS-CE):

\begin{equation}
 	 q_k(x)=\frac{ \lambda_k  e^{z_k(x)}}{\sum_{j=1}^{N_t} \lambda_j e^{z_j(x)} }  \;
 	 \text{ with } \lambda_i =
    \begin{cases}
        \epsilon,& \text{if } i\in P\\
        n_i, & \text{otherwise}
    \end{cases}
 \label{fig:relaxedBalancedSoftmax}
 \end{equation}
 where $\epsilon$ is usually a small positive number. 
 
While the expression of $\lambda_i$ for the new classes does not change, instead of using $\lambda_i$ equals to 0 for the old classes, the Relaxed Balanced Softmax Cross-Entropy uses a small positive value $\epsilon$ in order to slightly reduce the constraint on the model during the learning of the new classes. This new hyper-parameter $\epsilon$ can be adjusted to tune the balance between stability and plasticity. By using $\epsilon$ equal to 0.0, the Relaxed Balanced Softmax Cross-Entropy is equivalent to the Balanced Softmax Cross-Entropy as described in Equation~(\ref{eq:balancedSoftmax}).

\revisedsection{The selection of the value for $\epsilon$ remains an open question and is discussed in section \ref{section:epsilon}. Experimentally, the value of $\epsilon$ is fixed at a small percentage of the number of images per class in the training dataset.} \revisedsection{It should also be noted that the algorithm from Meta-Balanced Softmax Cross-Entropy can not be used in this setting as the balanced validation dataset that is necessary cannot be created in the zero memory setting.}

\section{Experiments}
\label{section:experiments}
\subsection{Experimental setups}

\subsubsection{Datasets}
Experiments are conducted on three competitive datasets for large-scale incremental learning: CIFAR100, ImageNet-Subset, and ImageNet. We use the experimental settings defined in~\citep{Hou_2019_CVPR} by initially training the models on the first half of the classes of the dataset (referred to as the base classes) during the base step before learning the remaining classes during the next 5, 10, or 25 incremental steps. \revisedsection{For class-incremental learning with memory,} a small growing memory for rehearsal that contains 20 samples per class for the old classes \revisedsection{is used}. Following~\citep{Hou_2019_CVPR, Rebuffi_2017_CVPR}, the class order is defined by NumPy using the random seed 1993.

\begin{itemize}
    \item \textbf{CIFAR100}~\citep{krizhevsky2009learning} is composed of 60,000 32x32 RGB images equally divided among 100 classes with 500 images for training and 100 for testing for each class. There are 50 base classes and the remaining 50 classes are learned in groups of 2, 5, or 10 classes depending on the number of incremental steps.
    \item \textbf{ImageNet} (ILSVRC 2012)~\citep{russakovsky2015ImageNet} is composed of about 1.3 million high-resolution RGB images divided among 1,000 classes with around 1,300 images for training and 50 for testing for each class. There are 500 base classes and the remaining 500 classes are learned in groups of 50 or 100 classes depending on the number of incremental steps.
    \item \textbf{ImageNet-Subset} is a subset of ImageNet only containing the first 100 classes. There are 50 base classes and the 50 remaining classes are learned in groups of 2, 5, or 10 classes depending on the number of incremental steps.
\end{itemize}

Additionally, we also conduct an experiment on ImageNet without a base step containing half of the dataset. This more challenging setting initially used in~\citep{Rebuffi_2017_CVPR,wu2019large} consists in learning the 1,000 classes of the ImageNet dataset in 10 training steps, each containing 100 new classes. For this setting, a fixed memory is used with a budget of 20,000 samples in total. 

\begin{table*}
   \caption{Average incremental accuracy (Top-1) on CIFAR100, ImageNet-Subset, and ImageNet with a base step containing half of the classes followed by 5, 10, and 25 incremental steps, using a growing memory of 20 samples per class for all methods. Results for iCaRL and LUCIR are reported from \citep{Hou_2019_CVPR} ; results for Mnemonics and BiC are reported from \citep{liu2020mnemonics}; results for PODNet, TPCIL, and DDE are reported from their respective paper. Results marked with \enquote{*} correspond to our own experiments. Results on CIFAR-100 averaged over 3 random runs. Results on ImageNet and ImageNet-Subset are reported as a single run. Best result is marked in bold and second best is underlined. Results without memory in Table~\ref{tab:averageAccNoMem}.}
\begin{center}
{\scriptsize
\begin{tabular}{lcccccccccc}
\toprule
 & \multicolumn{3}{c}{CIFAR100} & & \multicolumn{3}{c}{ImageNet-Subset} & & \multicolumn{2}{c}{ImageNet}  \\
\cline{2-4} \cline{6-8} \cline{10-11}
{\scriptsize Number of incremental steps} & 5 & 10 & 25 & & 5 & 10 & 25 & & 5 & 10 \\
\toprule
iCaRL~\citep{Rebuffi_2017_CVPR} & 57.17 & 52.57 & - & & 65.04 & 59.53 & - & & 51.36 & 46.72 \\
BiC~\citep{wu2019large} & 59.36 & 54.20 & 50.00 & & 70.07 & 64.96 & 57.73 & &  62.65 &  58.72 \\
LUCIR~\citep{Hou_2019_CVPR} & 63.42 & 60.18 & - & & 70.47 & 68.09 & - & & 64.34 & 61.28 \\
LUCIR w/ Mnemonics~\citep{liu2020mnemonics} & 63.34 & 62.28 & \underline{60.96} & & 72.58 & 71.37 & \textbf{69.74} & & 64.54 & 63.01 \\ 
LUCIR w/ DDE~\citep{Hu_2021_CVPR} & 65.27 & 62.36 & - & & 72.34 & 70.20 & - & & 67.51 & 65.77 \\
PODNet~\citep{douillard2020podnet} & 64.83 & 63.19 & 60.72 & & 75.54 & 74.33 & 68.31 & & 66.95 & 64.13 \\
PODNet w/ DDE~\citep{Hu_2021_CVPR} & \underline{65.42} & \underline{64.12} & - & & \textbf{76.71} & \textbf{75.41} & - & & 66.42 & 64.71 \\
TPCIL~\citep{10.1007/978-3-030-58529-7_16} & 65.34 & 63.58 & - & & \underline{76.27} & 74.81 & - & & 64.89 & 62.88 \\
\midrule 
IL-Baseline* & 43.80  & 37.00 & 30.57 & & 51.52 & 42.22 & 30.96 & & 43.23 & 36.70 \\
 IL-Baseline w/ BalancedS-CE (ours) & 62.22 & 58.32 & 52.10 & & 72.57 & 68.25 & 61.49 & & 66.45 & 62.14  \\
 IL-Baseline w/ Meta BalancedS-CE (ours) & 64.11 & 60.08 & 49.63 & & 72.88 & 69.26 & 60.70 & & 66.15 & 61.59  \\
\midrule
LUCIR~\citep{Hou_2019_CVPR}* & 63.37 & 60.88 & 57.07 & & 70.25 & 67.84 & 63.23 & & 66.69 & 64.06  \\
 LUCIR w/ BalancedS-CE (ours) & 64.83 & 62.36 & 58.38 & & 71.18 & 70.66 & 65.10 & & \underline{67.81}  &  \underline{66.47} \\
\midrule 
PODNet~\citep{douillard2020podnet}* & 64.46 & 62.69 & 60.63 & & 74.97 & 71.57 & 62.67 & & 65.20  & 62.87  \\
PODNet w/ BalancedS-CE (ours) & \textbf{67.77} & \textbf{66.57} & \textbf{64.64} & & 76.08 & \underline{74.93}  & \underline{69.46} & &  \textbf{69.67} &  \textbf{68.65} \\
\bottomrule
\end{tabular}}
\end{center}
\label{tab:averageAcc}
\end{table*}

\subsubsection{Baselines}
The IL-Baseline which uses the Standard Softmax Cross-Entropy loss is used to highlight the impact of the Balanced Softmax Cross-Entropy loss function for class incremental learning. This baseline is considered as the lower-bound method. 
Furthermore, the proposed models are compared with several recent approaches for class incremental learning with a replay memory: iCaRL~\citep{Rebuffi_2017_CVPR}, BiC~\citep{wu2019large}, LUCIR~\citep{Hou_2019_CVPR}, Mnemonics~\citep{liu2020mnemonics}, PODNet~\citep{douillard2020podnet}, TPCIL~\citep{10.1007/978-3-030-58529-7_16} and DDE~\citep{Hu_2021_CVPR}. For the class incremental learning without memory settings, \revisedsection{the Nearest Class Mean (NCM) classifier and the Streaming Linear Discriminant Analysis (SLDA)~\citep{1510767,Hayes_2020} are used in addition} to two recent approaches: DDE~\citep{Hu_2021_CVPR} and SPB~\citep{Wu_2021_ICCV} and its derivatives (SPB-I and SPB-M). \revisedsection{Contrary to the other considered approaches, both the NCM classifier and SLDA rely on a feature extractor which is only trained during the base step and not further finetuned during the following incremental steps. It results in quick to train methods that achieve competitive results, especially when having a large base step.}

To measure the performance of the different models and compare them, the average incremental accuracy is used as in~\cite{Rebuffi_2017_CVPR}. It is defined as the average of the Top-1 accuracy of the model on the test data \revisedsection{of all the classes seen so far} at the end of each training step, including the initial base step.

\subsubsection{Implementation details}

\begin{table}
   \caption{Average incremental accuracy (Top-1 and Top-5) on ImageNet with 10 training steps of 100 classes each using a fixed memory of 20,000 samples. Results for iCaRL, EEIL, and BiC are reported from \citep{wu2019large} ; results for SS-IL are reported from its respective paper. \revisedsection{Results marked with \enquote{*} correspond to our own experiments.} Results reported as a single run. Best result is marked in bold and second best is underlined.}
\begin{center}
{\scriptsize
\begin{tabularx}{\columnwidth}{bvv}
\toprule
 &  \multicolumn{2}{>{\centering\hsize=.5\hsize}X}{\makecell{ImageNet with \\ 10 training steps}}\\
\cline{2-3}
 & \makecell{Top-1 Avg.\\ Inc. Acc.} & \makecell{Top-5 Avg.\\ Inc. Acc.}    \\
\toprule
iCaRL~\citep{Rebuffi_2017_CVPR} & - & 63.7 \\
EEIL~\citep{10.1007/978-3-030-01258-8_15} & - & 72.2 \\
BiC~\citep{wu2019large} & - & 84.0\\
SS-IL~\citep{Ahn_2021_ICCV} & \textbf{65.2} & \textbf{86.7}\\
\midrule 
 \revisedsection{IL-Baseline*} & \revisedsection{46.65} & \revisedsection{68.74} \\
 IL-Baseline w/ BalancedS-CE (ours) & \underline{\revisedsection{64.15}} & \revisedsection{85.65} \\
 IL-Baseline w/ Meta BalancedS-CE (ours) & \revisedsection{63.57} & \revisedsection{85.43} \\
 \midrule 
 \revisedsection{LUCIR~\citep{Hou_2019_CVPR}*} & \revisedsection{60.40} & \revisedsection{82.31} \\
\revisedsection{LUCIR w/ BalancedS-CE (ours)} & \revisedsection{60.75} &  \revisedsection{84.31} \\
 \midrule 
  \revisedsection{PODNet~\citep{douillard2020podnet}*} & \revisedsection{62.58} & \revisedsection{83.50} \\
\revisedsection{PODNet w/ BalancedS-CE (ours)} & \revisedsection{64.08} & \underline{\revisedsection{85.71}} \\
\bottomrule
\end{tabularx}}
\end{center}
\label{tab:averageAccFromScratch}
\end{table}

All compared methods use the 32-layer ResNet~\citep{He_2016} for CIFAR100 and the 18-layers ResNet for ImageNet and ImageNet-Subset. In our experiments, the input images are normalized, randomly horizontally flipped, and cropped with no further augmentation applied.

For CIFAR100, the IL-Baseline model is trained for 250 epochs using SGD with momentum optimizer with a batch size of 128 and weight decay set to 0.0002. The learning rate starts at 0.1 and is divided by 10 after the epochs 100, 150, and 200.
For ImageNet and ImageNet-Subset, the IL-Baseline model is trained for 100 epochs using SGD with momentum optimizer with a batch size of 256 and weight decay set to 0.0001. The learning rate starts at 0.1 and is divided by 10 after the epochs 30, 60, 80, and 90.
For all datasets, the temperature $T$ of the distillation loss for the IL-Baseline is equal to 2.
When the Balanced Softmax Cross-Entropy loss or the Relaxed Balanced Softmax Cross-Entropy are combined with other approaches for class incremental learning, the same hyper-parameters and training procedure as those reported in their respective original publications are used. For all datasets, the weighting coefficient $\alpha$ of the Balanced Softmax Cross-Entropy is set to 1.0 \revisedsection{as it corresponds to Equation~(\ref{eq:balancedSoftmax})} and the value of $\epsilon$ for the Relaxed Balanced Softmax Cross-Entropy is set to 0.2\% of the number of training images per class, which corresponds to $\epsilon = 1$ for CIFAR-100 and $\epsilon = 2.6$ for ImageNet-Subset. For Meta-Balanced Softmax Cross-Entropy, 10\% of the memory size is used for the balanced validation set $B_t$, which represents 2 samples per class for the main experiments. In order to decrease the computational requirement of the method, the $\alpha$ weighting coefficient is only updated every 10 optimization steps instead of every optimization step. PyTorch has been used for the experiments and gradients through the optimization process are computed using Higher library~\citep{grefenstette2019generalized}.

\begin{table*}
   \caption{Average incremental accuracy (Top-1) on CIFAR100, ImageNet-Subset\revisedsection{, and ImageNet} with a base step containing half of the classes followed by 5 and 10 incremental steps without using memory for storing samples from past classes. Results for SPB, SPB-I, and SPB-M are reported from \citep{Wu_2021_ICCV} ; results for DDE are reported from \citep{Hu_2021_CVPR}. Results marked with \enquote{*} correspond to our own experiments. Results on CIFAR-100 and ImageNet-Subset averaged over 3 random runs, \revisedsection{results on ImageNet reported as a single run}. Best result is marked in bold and second best is underlined.}
\begin{center}
{\scriptsize
\begin{tabular}{lcccccccc}
\toprule
 & \multicolumn{2}{c}{CIFAR100} & & \multicolumn{2}{c}{ImageNet-Subset} & & \multicolumn{2}{c}{ImageNet}\\
\cline{2-3} \cline{5-6} \cline{8-9}
{\scriptsize Number of incremental steps} & 5 & 10 & & 5 & 10 & & 5 & 10\\
\toprule
SPB~\citep{Wu_2021_ICCV} & 60.9 & 60.4 & & 68.7 & 67.2 & & \revisedsection{57.8} & \revisedsection{-} \\
SPB-I~\citep{Wu_2021_ICCV} & 62.6 & 62.7 & & 70.1 & 69.8 & & \revisedsection{58.0} & \revisedsection{-} \\
SPB-M~\citep{Wu_2021_ICCV} & \underline{65.5} & \textbf{65.2} & & \underline{71.7} & \underline{70.6} & & \revisedsection{59.7} & \revisedsection{-} \\
DDE~\citep{Hu_2021_CVPR} & 59.11 & 55.31 & & 69.22 & 65.51 & & \revisedsection{-} & \revisedsection{-} \\
\midrule
\revisedsection{NCM classifier*} & \revisedsection{59.80} &  \revisedsection{59.52} & & \revisedsection{65.91} &  \revisedsection{65.71} && \revisedsection{59.92} & \revisedsection{59.59} \\
\revisedsection{SLDA~\citep{hayes2019lifelong}*} & \revisedsection{60.20} &  \revisedsection{60.13} & & \revisedsection{67.07} &  \revisedsection{66.84} && \revisedsection{61.69} & \revisedsection{61.47} \\
\midrule 
LUCIR~\citep{Hou_2019_CVPR}* & 43.71 & 32.08 & & 60.70 & 47.97 & & \revisedsection{51.19} & \revisedsection{39.72} \\
LUCIR w/ BalancedS-CE (ours) & 54.68 & 50.87 & & 58.63 & 57.71 & & \revisedsection{63.29} & \revisedsection{62.90} \\
LUCIR w/ Relaxed BalancedS-CE (ours) & 59.43 & 54.93 & & 66.27 & 65.12 & & \revisedsection{ \underline{67.07}} & \revisedsection{\underline{65.07}} \\
\midrule 
PODNet~\citep{douillard2020podnet}* & 41.85 & 35.87 & & 66.61 & 55.14 & & \revisedsection{47.65} & \revisedsection{37.22} \\
PODNet w/ BalancedS-CE (ours) & 63.31 & 62.39 & & 63.94 & 61.98 & & \revisedsection{55.00} & \revisedsection{54.62} \\
PODNet w/ Relaxed BalancedS-CE (ours) & \textbf{65.66} & \underline{63.82} & & \textbf{72.56} & \textbf{70.99} & & \revisedsection{\textbf{68.50}} & \revisedsection{\textbf{67.38}} \\
\bottomrule
\end{tabular}}
\end{center}
\label{tab:averageAccNoMem}
\end{table*}

\subsection{Comparison Results}

\subsubsection{Class-incremental Learning with memory}

The average incremental accuracy (Top-1) on CIFAR100, ImageNet-Subset, and ImageNet of our methods and the different baselines are reported in Table~\ref{tab:averageAcc} using 5, 10, and 25 incremental steps settings. For a fair comparison, each method uses a growing memory containing exactly 20 examples per class.

First, we use the IL-Baseline to support the importance of bias mitigation for large-scale incremental learning. This approach does not use any bias mitigation method and thus highlights the difference between the Standard Softmax Cross-Entropy and the Balanced Softmax Cross-Entropy. On every dataset and in every setting, using the Balanced Softmax Cross-Entropy to train the IL-Baseline instead of the Standard Softmax Cross-Entropy loss results in a huge increase of the average incremental accuracy. Moreover, by using the meta-learning algorithm to learn the weighting coefficient $\alpha$ instead of using the fixed value of 1.0 it is possible to further improve the average incremental accuracy of the IL-Baseline. The largest improvement appears on CIFAR100 with 5 and 10 incremental steps. However, on ImageNet-Subset and ImageNet, there are no significant advantage for the Meta Balanced Softmax Cross-Entropy over the Balanced Softmax Cross-Entropy, and in some cases the Meta Balanced Softmax Cross-Entropy performs worse. On the demanding 25 incremental steps settings, even though the Meta Balanced Softmax Cross-Entropy achieves better performance than the Balanced Softmax Cross-Entropy during the first few incremental steps, its performance deteriorates in the long term and achieves a lower average incremental accuracy at the end. Overall, the IL-Baseline trained with the Balanced Softmax Cross-Entropy achieves superior average incremental accuracy than BiC~\citep{wu2019large} and similar performances to LUCIR~\citep{Hou_2019_CVPR}.

To demonstrate the flexibility of the proposed method, the Balanced Softmax Cross-Entropy loss is combined with two state-of-the-art approaches: LUCIR~\citep{Hou_2019_CVPR} and PODNet~\citep{douillard2020podnet}. For both methods, the classification loss previously used, the Standard Softmax Cross-Entropy loss in the case of LUCIR and the NCA loss~\citep{NIPS2004_42fe8808, Movshovitz_Attias_2017} in the case of PODNet, is replaced by the Balanced Softmax Cross-entropy \revisedsection{and the balanced finetuning step is removed} without any other modifications to the method or its hyper-parameters. On every dataset and in every setting, using the Balanced Softmax Cross-Entropy significantly improves the average incremental accuracy of both LUCIR and PODNet. It also results in a decrease of the computational complexity by removing the necessity of performing an additional finetuning step on a small balanced dataset after each incremental step, as previously done by both methods in some settings. In addition to reducing the training time, removing the finetuning step also means that it is now possible to perform inference at any time during the training procedure. The Balanced Softmax Cross-Entropy loss increases the average incremental accuracy from 0.93 up to 2.81 percentage points (\emph{p.p.}) for LUCIR and from 1.11 up to 6.79 percentage points for PODNet. On ImageNet with 5 incremental steps, PODNet combined with the Balanced Softmax Cross-Entropy outperforms PODNet by 4.47 \emph{p.p.}\ in terms of average incremental accuracy and reaches final overall accuracy of 64.4\% which is about 6 \emph{p.p.}\ below the theoretical Top-1 accuracy of a model trained on the whole dataset at once. It is important to note that our proposed method is complementary to other approaches such as Mnemonics~\citep{liu2020mnemonics}, DDE~\citep{Hu_2021_CVPR} or GeoDL~\citep{Simon_2021_CVPR}, and they could be combined in order to further improve the performance.

Additionally, Table~\ref{tab:averageAccFromScratch} contains the average incremental accuracy (Top-1 and Top-5) of our proposed methods and the baselines on ImageNet using 10 training steps each containing 100 new classes as used by \cite{Rebuffi_2017_CVPR} and \cite{wu2019large}. \revisedsection{By replacing the standard Softmax Cross-Entropy with the Balanced Softmax Cross-Entropy, the Top-1 average incremental accuracy of the IL-Baseline increased from 46.65\% to 64.15\% and the Top-5 average incremental accuracy increased from 68.74\% to 85.65\% without increasing the training time.} Comparing BiC~\citep{wu2019large} with the IL-Baseline combined with Balanced Softmax Cross-Entropy highlights the impact of the proposed Balanced Softmax Cross-entropy for incremental learning, as the only difference between the two models is the bias mitigation method used. Our proposed method achieves a Top-5 average incremental accuracy of \revisedsection{85.65\%} which represents a notable improvement compared to BiC while not requiring an additional training step to learn the bias correction parameters at the end of each incremental step. The proposed method also achieves competitive results with the recently proposed SS-IL~\citep{Ahn_2021_ICCV}, without using neither the ratio-preserving mini-batch nor the improved distillation loss. The meta-learning approach does not improve the performance of the model in this particular setting, though. \revisedsection{In this setting also, the proposed Balanced Softmax Cross-Entropy can be combined with LUCIR and PODNet to improve their performance while decreasing their training time and allowing inference at any time by removing the balanced finetuning step. Even though the difference of TOP-1 average incremental accuracy between standard LUCIR and LUCIR combined with our proposed method seems to be limited in this particular setting, it is important to note that LUCIR with the Balanced Softmax achieves a significantly higher final TOP-1 overall accuracy: 51.41\% compared to the 48.30\% of the standard LUCIR. Moreover, by removing the balanced finetuning step from the standard LUCIR to match the flexibility of our method, its TOP-1 average incremental accuracy drops to 56.41\% and its final TOP-1 overall accuracy also drops to 40.44\%.}

\subsubsection{Class-incremental Learning without memory}
Next, we compare our method and the different baselines for class-incremental learning without memory using both 5 and 10 incremental steps.
The average incremental accuracy (Top-1) on CIFAR100, ImageNet-Subset, \revisedsection{and ImageNet} are reported in Table~\ref{tab:averageAccNoMem}.

Even though \revisedsection{in some cases}, using the Balanced Softmax Cross-Entropy loss may improve the average incremental accuracy of methods such as LUCIR and PODNet, it is not a viable option for the no memory setting as it tends to over-favor the old classes to the detriment of the new ones. However, by combining LUCIR with the proposed Relaxed Balanced Softmax Cross-Entropy, it is possible to achieve competitive performances compared to DDE\citep{Hu_2021_CVPR} without increasing the computational complexity of the training procedure compared to the standard LUCIR. Likewise, by combining PODNet with the proposed Relaxed Balanced Softmax Cross-Entropy, it is possible to outperform SPB~\citep{Wu_2021_ICCV} and achieve similar performance to SPB-M~\citep{Wu_2021_ICCV} without using multiple perspectives or self-supervised learning. It should be noted that PODNet combined with the Relaxed Balanced Softmax Cross-Entropy without memory achieves comparable average incremental accuracy against standard PODNet with 20 samples per class in memory. \revisedsection{Moreover, on ImageNet, the most challenging dataset, LUCIR and PODNet combined with the proposed Relaxed Balanced Softmax Cross-Entropy can outperform the standard LUCIR and PODNet with 20 samples per class in memory by a significant margin.} 

As detailed in Section~\ref{section:epsilon}, by changing the $\epsilon$ parameter of the Relaxed Balanced Softmax Cross-Entropy, it is possible to slightly increase the average incremental accuracy. For example, PODNet with the Relaxed Balanced Softmax Cross-Entropy can reach up to 74.4\% on ImageNet-Subset with 5 incremental steps.

\subsection{Ablation study}
\subsubsection{Impact of the memory size}
The average incremental accuracy of the IL-Baseline on CIFAR100 with 5 incremental steps is reported in Table~\ref{tab:memory} for various numbers of samples per class stored in the memory depending on the loss function used for the training. The results for the Meta Balanced Softmax Cross-Entropy are only reported for memory size larger than one as this method requires a distinct validation set and training set. The difference between the Standard Softmax Cross-Entropy and the Balanced Softmax Cross-Entropy increases as the size of the memory decreases. With only one sample per class in the memory, the Balanced Softmax Cross-Entropy reaches almost the same average incremental accuracy as the Standard Softmax Cross-Entropy with 50 samples per class. The Meta Balanced Softmax Cross-Entropy further improves the performance and appears to be especially efficient in scenarios with highly restricted memory.

\begin{table}
   \caption{Average incremental accuracy (Top-1) on the test set of CIFAR100 with a base step containing half of the classes followed by 5 incremental steps of the Incremental Learning Baseline depending on the number of samples stored in memory for each class and the training loss. Results averaged over 3 random runs. } 
   \label{tab:memory}
{\scriptsize
\begin{tabularx}{\columnwidth}{bsssss}
\toprule
 & \multicolumn{5}{c}{Memory size} \\
\cline{2-6}
 &  1 & 5 & 10 & 20 & 50 \\
\toprule
 IL-Baseline & 24.83 & 30.59 & 37.27 & 43.80 & 52.99 \\
\midrule 
 IL-Baseline w/ BalancedS-CE & \textbf{51.55} & 56.93 & 60.11 & 62.22 & 64.44\\
 IL-Baseline w/ Meta BalancedS-CE & - & \textbf{62.13} & \textbf{63.02} & \textbf{64.11} & \textbf{65.60}  \\
\bottomrule
\end{tabularx}}
\end{table}

\begin{table}
   \caption{Accuracy on the test set of CIFAR100 with a base step containing half of the classes followed by \subref{tab:alpha:5} 5 and \subref{tab:alpha:10} 10 incremental steps of the Incremental Learning Baseline (IL-Baseline) depending on the value used for the weighing coefficient $\alpha$ of the Balanced Softmax Cross-Entropy; using a growing memory of 20 samples per class. Results averaged over 3 random runs. }
\begin{subtable}{\linewidth}
\begin{center}
\caption{CIFAR100 - 5 incremental steps}\label{tab:alpha:5}
{\scriptsize
\begin{tabularx}{\columnwidth}{bttt}
\toprule
 & \makecell{final\\base\\accuracy} & \makecell{final\\overall\\accuracy} & \makecell{average\\inc.\\accuracy} \\
\toprule
BalancedS-CE $\alpha=0.1$ & \textbf{68.38} & 51.54 & 62.68 \\
BalancedS-CE $\alpha=0.25$ & 63.99 & 54.88 & 64.09 \\
BalancedS-CE $\alpha=0.5$ & 58.78 & \textbf{55.44} &  63.80 \\
BalancedS-CE $\alpha=1.0$ & 52.08 & 54.55 &  62.22 \\
\midrule
Meta BalancedS-CE  & 61.20 & 55.21 & \textbf{64.11}\\
\bottomrule
\end{tabularx}}
\end{center}
\end{subtable}%

\vspace{0.5cm}

\begin{subtable}{\linewidth}
\begin{center}
\caption{CIFAR100 - 10 incremental steps}\label{tab:alpha:10}
{\scriptsize
\begin{tabularx}{\columnwidth}{bttt}
\toprule
 & \makecell{final\\base\\accuracy} & \makecell{final\\overall\\accuracy} & \makecell{average\\inc.\\accuracy} \\
\toprule
BalancedS-CE $\alpha=0.1$ & \textbf{63.43} & 44.71 & 57.68  \\
BalancedS-CE $\alpha=0.25$ & 60.28 & 47.86 & 59.40 \\
BalancedS-CE $\alpha=0.5$ & 56.48 & 49.62 & 59.52 \\
BalancedS-CE $\alpha=1.0$ & 51.01 & 49.57 & 58.32 \\
\midrule
Meta BalancedS-CE & 60.39 & \textbf{49.65} & \textbf{60.08} \\
\bottomrule
\end{tabularx}}
\end{center}
\end{subtable}
\label{tab:alpha}
\end{table}

\begin{figure*}
\begin{subfigure}[t]{0.475\linewidth}
        \begin{center}
        \includegraphics[width=\columnwidth]{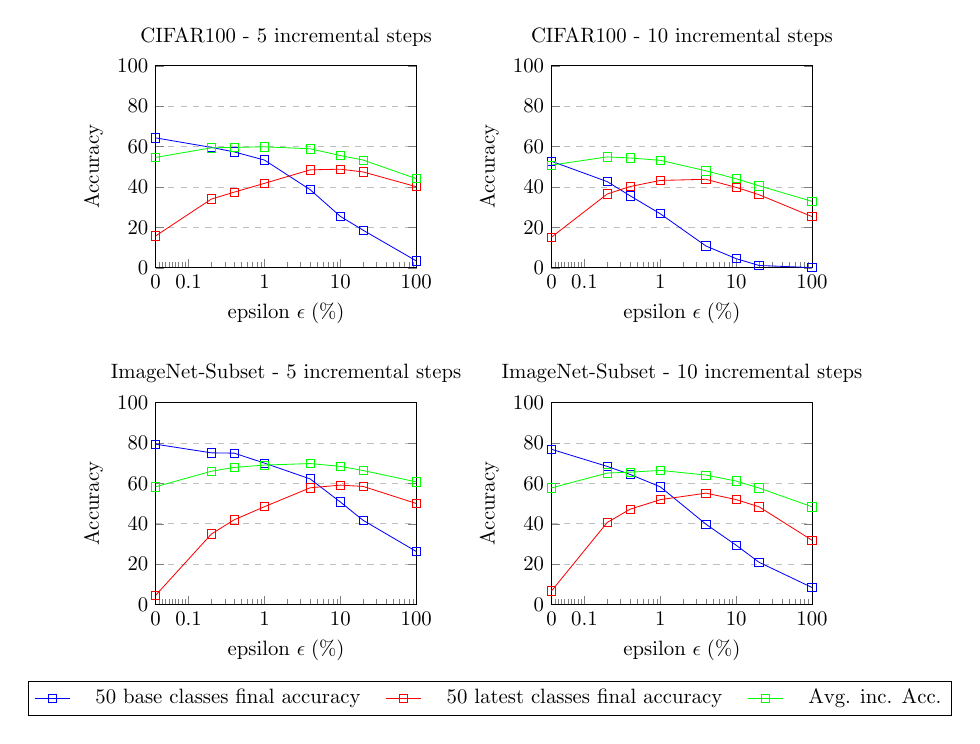}
        \caption{LUCIR /w Relaxed Balanced Softmax Cross-Entropy}
        \label{fig:relaxed_epsilon:lucir}
        \end{center}
    \end{subfigure}  \hfill
    \begin{subfigure}[t]{0.475\linewidth}
        \begin{center}
        \includegraphics[width=\columnwidth]{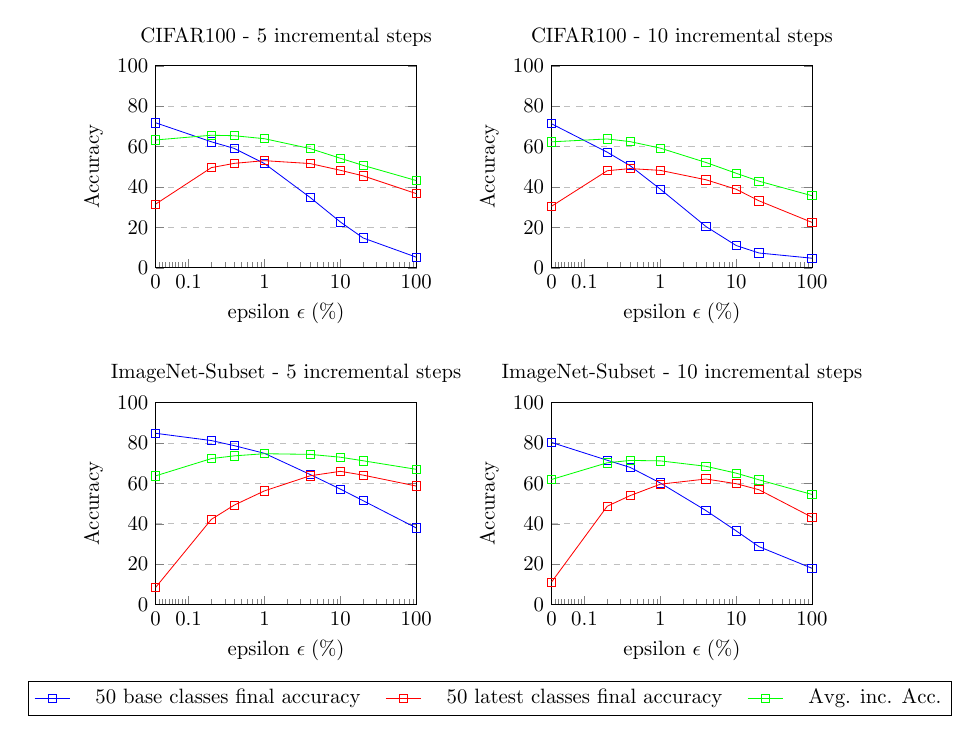}
        \caption{PODNet /w Relaxed Balanced Softmax Cross-Entropy}
        \label{fig:relaxed_epsilon:podnet}
        \end{center}
    \end{subfigure}
   \caption{Performance of \subref{fig:relaxed_epsilon:lucir} LUCIR and \subref{fig:relaxed_epsilon:podnet} PODNet combined with the Relaxed Balanced Softmax Cross-Entropy for different values of $\epsilon$ on CIFAR100 and ImageNet-Subset without memory. Parameter $\epsilon$ is expressed in percentage of the number of samples per class in the training dataset. X-axis represented using linear scale on [0,0.1] and logarithmic scale on [0.1, 100] for better visibility. Results on CIFAR-100 averaged over 3 random runs and results on ImageNet-Subset reported as a single run. Figure best viewed in color.}
\label{fig:relaxed_epsilon}
\end{figure*}

\subsubsection{Impact of the weighting coefficient alpha}
The average incremental accuracy of the IL-Baseline on CIFAR100 is reported in Table~\ref{tab:alpha} with \subref{tab:alpha:5} 5 incremental steps and \subref{tab:alpha:10} 10 incremental steps depending on the value of the weighting coefficient $\alpha$ used.

As expected, we can see in both settings that decreasing the value of the weighting coefficient $\alpha$ will reduce the plasticity of the model and thus increase the final accuracy of the 50 base classes. The smaller the weighting coefficient $\alpha$, the higher the final accuracy of the base classes. Depending on the constraints of the problem, it is possible to use the weighting coefficient $\alpha$ to balance the trade-off between rigidity and plasticity while limiting the impact on the overall final performance of the model: for example on CIFAR-100 with 5 incremental steps by using $\alpha = 0.1$, the accuracy of the 50 base classes increases by 16.30 \emph{p.p.}\ while the final overall accuracy of the model only decreases by 3.01 \emph{p.p.}\ compared to the default value of 1.0 for the Balanced Softmax Cross-Entropy.
Experimentally, it appears that using the default value of $\alpha = 1.0$ achieves the most balanced performances between the base classes and the latest classes in both the 5 and 10 incremental steps settings. However, it does not achieve the highest average incremental accuracy, and using a smaller value for $\alpha$ may improve it. To this end, using the proposed Meta Balanced Softmax Cross-Entropy is an effective solution as it reaches the highest average incremental accuracy in both settings. This demonstrates the advantage of this method for determining a correct value for the weighting coefficient $\alpha$ without performing several trials. However, the Meta Balanced Softmax Cross-Entropy seems to favor old classes. This may explain why this method does not improve the accuracy compared to the Balanced Softmax cross-entropy in cases where the latter already favors old classes as it is the case when combined with LUCIR for example.

\subsubsection{Impact of the hyper-parameter epsilon}
\label{section:epsilon}
Figure~\ref{fig:relaxed_epsilon} shows the performance of LUCIR and PODNet in the class incremental learning without memory setting when combined with the Relaxed Balanced Softmax Cross-Entropy depending on the value of $\epsilon$. For $\epsilon$ equals to 0\% of the number of samples per class in the training dataset, the Relaxed Balanced Softmax Cross-Entropy is equivalent to the Balanced Softmax Cross Entropy and for $\epsilon$ equals to 100\% of the number of samples per class in the training dataset it is almost equivalent to the Standard Softmax Cross Entropy.

As expected, increasing the value of $\epsilon$ for the Relaxed Balanced Softmax Cross-Entropy increases the plasticity of the model, the final accuracy of the 50 base classes decreases while the final accuracy of the 50 latest classes increases. This continues until a point where the final accuracy of the latest classes also decreases as the model is only able to retain the last few classes from the last incremental steps. By replacing the Balanced Softmax Cross-Entropy with the Relaxed Balanced Softmax Cross-Entropy using the default value of $\epsilon$, the final accuracy on the 50 latest classes can be drastically increased while limiting the impact on the 50 base classes: for example on ImageNet-Subset with 5 incremental steps using LUCIR, the final accuracy on the latest classes is increased by 30.40 \emph{p.p.}\ while the final accuracy on the base classes is only decreased by 4.28 \emph{p.p.}\ . 

\revisedsection{As epsilon is comparable to a number of images, the default value was chosen to be equal to 1 on CIFAR-100, the smallest strictly positive integer. This corresponds to 0.2\% of the number of images per class in the training dataset. It should be noted that the value of zero is equivalent to the \enquote{non-relaxed} Balanced Softmax Cross-Entropy. Experiments in Table~\ref{tab:averageAccNoMem} empirically show that this default value achieves competitive accuracy and can be generalized to all considered datasets regardless of the number of incremental steps or the approach it is combined with. Although manual tuning is not required to achieve high performance,} doing so can slightly increase the average incremental accuracy. For example, by using $\epsilon$ equals to 1.0\% of the number of training images per class for ImageNet-Subset, the average incremental accuracy can be increased from 0.94 \emph{p.p.}\ to 3.05 \emph{p.p.}\ depending on the method and the number of incremental steps. However, \revisedsection{even though Figure~\ref{fig:relaxed_epsilon} shows that there is a large range of $\epsilon$ values achieving similar competitive performance,} determining the optimal value of $\epsilon$ beforehand remains an open challenge and will be further investigated in future works. Experimental results show that the value of $\epsilon$ achieving either the highest accuracy or the best balance between rigidity and plasticity after the first incremental step is not the value that will achieve the best average incremental accuracy, the best final accuracy, or the best balance at the end of the complete training. This suggests that modifying the value of $\epsilon$ during the training procedure instead of using a fixed value may further improve the performance of the proposed approach.

Experimentally, it appears that by using a specific value of $\epsilon$, the Relaxed Balanced Softmax Cross-Entropy can perfectly balance the trade-off between rigidity and plasticity in all settings. This is shown by the Relaxed Balanced Softmax Cross-Entropy achieving the same accuracy for both the 50 base classes and the 50 latest classes learned during the following 5 or 10 incremental steps. Moreover, the highest average incremental accuracy occurs at or in the neighborhood of the value of $\epsilon$ achieving the best balance between rigidity and plasticity at the end of the training procedure.

\subsubsection{Mitigation of imbalance}

\begin{table}
   \caption{Average incremental accuracy (Top-1) on the test set of CIFAR100 with a base step containing half of the classes followed by 5 incremental steps of the Incremental Learning Baseline depending on the bias correction method used; using a growing memory of 20 samples per class. Results averaged over 3 random runs. } \label{tab:lossfunct}

{\scriptsize
\begin{tabularx}{\columnwidth}{bu}
\toprule
& \makecell{average\\incremental\\accuracy} \\
\toprule
IL-Baseline & 43.80  \\
\midrule
IL-Baseline w/ memory oversampling & 49.75 \\
IL-Baseline w/ class oversampling  & 55.95  \\
IL-Baseline w/ loss rescaling & 57.01  \\
IL-Baseline w/ balanced finetuning & 59.46 \\
IL-Baseline w/ Separated Softmax~\citep{Ahn_2021_ICCV} + oversampling & 61.21 \\
\midrule
IL-Baseline w/ BalancedS-CE (ours) &  62.22 \\
IL-Baseline w/ Meta BalancedS-CE (ours) & \textbf{64.11}\\
\bottomrule
\end{tabularx}}
\label{tab:imbalanceMitigation}
\end{table}

In Table ~\ref{tab:lossfunct}, different bias correction methods are compared with the Balanced Softmax Cross-Entropy and the Meta Balanced Softmax Cross-Entropy on CIFAR100 with 5 incremental steps. Both the IL-Baseline with memory oversampling and the IL-Baseline with class oversampling correspond to the IL-Baseline combined with a form of oversampling on the replay memory. Memory oversampling ensures that each training mini-batch theoretically contains the same number of samples from new and old classes while class oversampling ensures that each class has the same probability of appearing in each training mini-batch. The IL-Baseline with loss rescaling corresponds to the IL-Baseline where the standard Softmax Cross-Entropy loss for each sample is scaled inversely proportionally to the number of samples for the corresponding label in the training dataset. The IL-Baseline with balanced finetuning corresponds to the IL-Baseline being finetuned after each incremental step on a small balanced training set for few epochs similar to the procedure proposed by \cite{10.1007/978-3-030-01258-8_15} and used by PODNet and LUCIR. IL-Baseline with Separated Softmax relies on oversampling of the replay memory combined with the Separated Softmax layer~\citep{Ahn_2021_ICCV}.

In the experiments, it appears that using oversampling for the old classes is a simple yet effective method for mitigating the imbalance occurring in large scale incremental learning scenarios. However, this approach increases the total number of mini-batches resulting in a larger computational complexity of the training procedure and may also be prone to overfitting on the old classes. By rescaling the Softmax Cross-Entropy loss, it is possible to further increase the accuracy of the IL-Baseline without relying on oversampling. Finally, finetuning the model afterward on a small balanced dataset, as it is done by several state-of-the-art methods, further improves the final overall accuracy of the model at the cost of an additional training step after each incremental step. Our two proposed methods achieve higher average incremental accuracy than the other bias correction methods without requiring a two steps training procedure or oversampling.

\begin{figure}[ht]
\begin{subfigure}{\columnwidth}
        \centering
        \includegraphics[width=0.75\columnwidth]{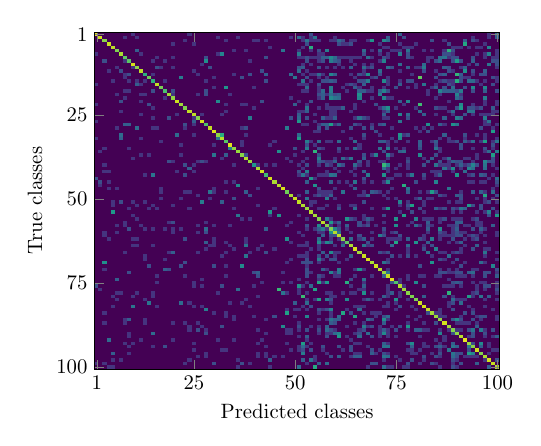}
        \caption{IL-Baseline w/ Balanced Softmax Cross-Entropy \\ Final overall top-1 accuracy: 54.76\%}
        \label{fig:confmat_ILbaseline_combined:balsoft}
    \end{subfigure}
    
    \vspace{\floatsep}
    
    \begin{subfigure}{\columnwidth}
        \centering
        \includegraphics[width=0.75\columnwidth]{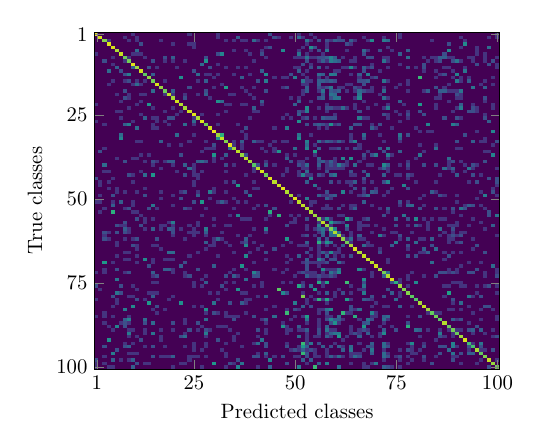} 
        \caption{IL-Baseline w/ Meta-Balanced Softmax Cross-Entropy \\ Final overall top-1 accuracy: 55.35\%}
        \label{fig:confmat_ILbaseline_combined:metabalsoft}
    \end{subfigure}
   \caption{Confusion matrix (transformed by log(1 + x) for better visibility) of the IL-Baseline trained using \subref{fig:confmat_ILbaseline_combined:balsoft} the Balanced Softmax Cross-Entropy and \subref{fig:confmat_ILbaseline_combined:metabalsoft} the Meta-Balanced Softmax Cross-Entropy on CIFAR100 with a base step containing half of the classes followed by 5 incremental steps, using 20 sample per class stored in the replay memory. Figure best viewed in color.}
\label{fig:confmat_ILbaseline_combined}
\end{figure}

Figure \ref{fig:confmat_ILbaseline_combined} shows the confusion matrices for IL-Baseline trained on CIFAR100 with 5 incremental steps using 20 sample per class stored in the replay memory when trained with \subref{fig:confmat_ILbaseline_combined:balsoft} the Balanced Softmax Cross-Entropy and \subref{fig:confmat_ILbaseline_combined:metabalsoft} the Meta Balanced Softmax Cross-Entropy. Visual comparison with Figure \ref{fig:confmat_ILbaseline} depicting the IL-Baseline trained using the Standard Softmax Cross-Entropy in the same setting confirms that our proposed method is efficient in mitigating the imbalance between the new and old classes in the class incremental learning scenario.

\section{Conclusion}
\label{section:conclusion}

In our work, following recent advances in Long-Tailed Visual Recognition, we proposed to use the Balanced Softmax Cross-Entropy loss function as a straightforward replacement for the Standard Softmax Cross-Entropy loss in order to correct the bias towards new classes in class incremental learning with a replay memory. The expression of this loss offers control on the plasticity-rigidity trade-off of the model but also on the importance of each class individually. We further investigated how it could be adapted for the challenging class incremental learning without memory setting and proposed an extension called the Relaxed Balanced Softmax Cross-Entropy.

Experiments on CIFAR100, ImageNet-Subset, and ImageNet with various settings showed that by seamlessly combining our proposed method with state-of-the-art approaches for incremental learning, it is possible to further increase the accuracy of those methods while also potentially decreasing the computational cost of the training procedure by removing the need for a balanced finetuning step. 

In our future work, we will explore how the coefficients of the Balanced Softmax Cross-Entropy can be tuned during the training procedure in order to improve the accuracy of the model and the balance between rigidity and plasticity when combined with memory and when used without memory. \revisedsection{The use of the Balanced Softmax Cross-Entropy for general incremental learning on imbalanced datasets is also left for our future work.}

\section*{Declaration of Competing Interest}
The authors declare that they have no known competing financial interests or personal relationships that could have appeared to influence the work reported in this paper.

\section*{Acknowledgment}
This work is partly supported by JST CREST (Grant Number JPMJCR1687), JSPS Grant-in-Aid for Scientific Research (Grant Number 17H01785, 21K12042), and the New Energy and Industrial Technology Development Organization (Grant Number JPNP20006).

\bibliographystyle{model2-names}
\bibliography{bib.bib}

\end{document}